# Designing Complex Experiments
# by Applying Unsupervised Machine Learning

**Alex Glushkovsky**

BMO Financial Group

**Abstract**

Design of experiments (DOE) is playing an essential role in learning and improving a variety of objects and processes. The article discusses the application of unsupervised machine learning to support the pragmatic designs of complex experiments. Complex experiments are characterized by having a large number of factors, mixed-level designs, and may be subject to constraints that eliminate some unfeasible trials for various reasons. Having such attributes, it is very challenging to design pragmatic experiments that are economically, operationally, and timely sound. It means a significant decrease in the number of required trials from a full factorial design, while still attempting to achieve the defined objectives. A beta variational autoencoder (beta-VAE) has been applied to represent trials of the initial full factorial design after filtering out unfeasible trials on the low dimensional latent space. Regarding visualization and interpretability, the paper is limited to 2D representations. Beta-VAE supports (1) orthogonality of the latent space dimensions, (2) isotropic multivariate standard normal distribution of the representation on the latent space, (3) disentanglement of the latent space representation by levels of factors, (4) propagation of the applied constraints of the initial design into the latent space, and (5) generation of trials by decoding latent space points. Having an initial design representation on the latent space with such properties, it allows for the generation of pragmatic design of experiments (G-DOE) by specifying the number of trials and their pattern on the latent space, such as square or polar grids. Clustering and aggregated gradient metrics have been shown to guide grid specification.

## 1       Introduction

Design of experiments (DOE) is a methodology of learning by actively changing factors that assumably have impacts on the response of interest of an object or a process. It is backed by analytical principles, statistical analysis, and modeling.

When designing experiments, there are two major questions typically in focus:

- What are the optimal levels of factors with respect to the defined response (also known as the objective function or the target variable) that has a certain goal: the higher, the better; the lower, the better; or the target value, the better?
- How important are factors with respect to their impact on the defined response?

The challenge of answering the above-mentioned questions is that it may require many trials to collect sufficient information especially when dealing with complex experiments. In the context of this paper, the complex experiments are categorized as having a large number of factors, mixed-level designs, and may be subject to constraints that forbid some trials. The latest may be caused by safety, regulatory, physical, operational, legal, ethical, economical, or other reasons.

There are many methodologies supporting design of experiments, such as full factorial, fractional factorial (Box *et al*, 1978; Montgomery, 2012), screening (Plackett and Burman, 1946), Taguchi L-designs (Ross, 1996), response surface methodology (Box and Wilson, 1951), dual response (Vining and Myers, 1990; Castillo and Montgomery, 1993; Lin and Tu, 1995), and Nelder–Mead technique (Nelder and Mead, 1965).

Suitable design can be established by defining the number of factors, number of levels, and desired resolution; and then querying against numerous published matrixes. However, when dealing with complex experiments, it is still very challenging to come up with the appropriate pragmatic design.

Essentially, design of experiments addresses the efficiency of learning by changing various factors.

Another learning mechanism is machine learning. Machine learning is dealing with already collected data or by repeatedly collecting data under some stochastic variations (i.e., reinforcement learning) while gradually achieving the defined objectives.

Machine learning and DOE both have a statistical background and are mutually supportive. Thus, supervised machine learning is used for backend modeling of the obtained results in some experiments. On the other hand, design of experiments is widely utilized for tuning of hyperparameters in machine learning (for example, Koch *et al*, 2107; Glushkovsky, 2018; Lujan-Moreno *et al*, 2018; Campbell *et al*, 2021). The tuning of hyperparameters even becomes an embedded part of some machine learning algorithms.

Current unprecedented development in machine learning triggers the intriguing question: "Could machine learning support the design of complex experiments?" More specifically: "Is it possible to have a tool that supports the generation of a defined number of trials while controlling the orthogonality, balance, and meaningful coverage of the space defined by the full factorial design that has been subject to applicable constraints?" Reflecting on three key words: "generation", "orthogonality", and "space" in the question above leads to the consideration of generative autoencoders. These unsupervised machine learning algorithms have the following features: they represent information of input data on a low-dimensional latent space, disentangle underlying characteristics of the input objects there, and generate a decoded output for a given location on the latent space.

Recently, numerous great approaches have been developed that address representation on the latent space. It includes embedding and variational autoencoders, such as t-SNE, β-VAE, InfoVAE, InfoGAN (Van der Maaten and Geoffrey, 2008; Bengio, 2013; Bengio *et al*, 2013; Kingma and Welling, 2014; Chen *et al*, 2016; Makhzani *et al*, 2016; Higgins *et al*, 2017; Jang *et a*l, 2017; Dupont, 2018; Zhao *et al*, 2018, Rol´ınek, 2019).

In this paper, generative β-VAE has been applied considering the regularization term of Kullback–Leibler divergence from standard isotropic multivariate normal distribution. The beta parameter, that is a Lagrangian multiplier, provides flexibility to balance between decoded accuracy and disentanglement (Higgins *et al*, 2017).

The paper discusses experiences of unsupervised training of β-VAE to represent an initial full factorial design after satisfying all constraints on the 2D latent space forcing disentanglement and isotropic normal distribution. Having such representation, pragmatic design can be generated (G-DOE) by defining the number of trials and specifying the grid pattern. Essentially, it means a systematic and controllable sampling of points on the latent space.

## 2     Process Flow when Designing Complex Experiments by Applying β-VAE

When planning any experiment, it starts with the formalisation of objectives, including the definition of response of interest, followed by the selection of factors to be included and their levels, and then the specification of constraints which avert some trials, if applicable. Having that fulfilled, the generation of the initial DOE is a straightforward step: it is a full factorial design (i.e., a matrix of cartesian products between all factors) that has been subject to the exclusion of some trials that violate constraints.

Considering complex experiments, the initial design requires a large number of trials and usually cannot be realized due to economical, operational, or timing barriers. Even though there are numerous books, publications, tables, and software available that provide designs, it is very challenging to define pragmatic experiments, especially for mixed and/or constrained designs.

The application of β-VAE provides (1) the representation of initial DOE on the latent space, and (2) the generation of the pragmatic design as decoding from the specified grid on the latent space.

The process flow of designing complex yet pragmatic experiments by applying β-VAE is presented in Figure 1.

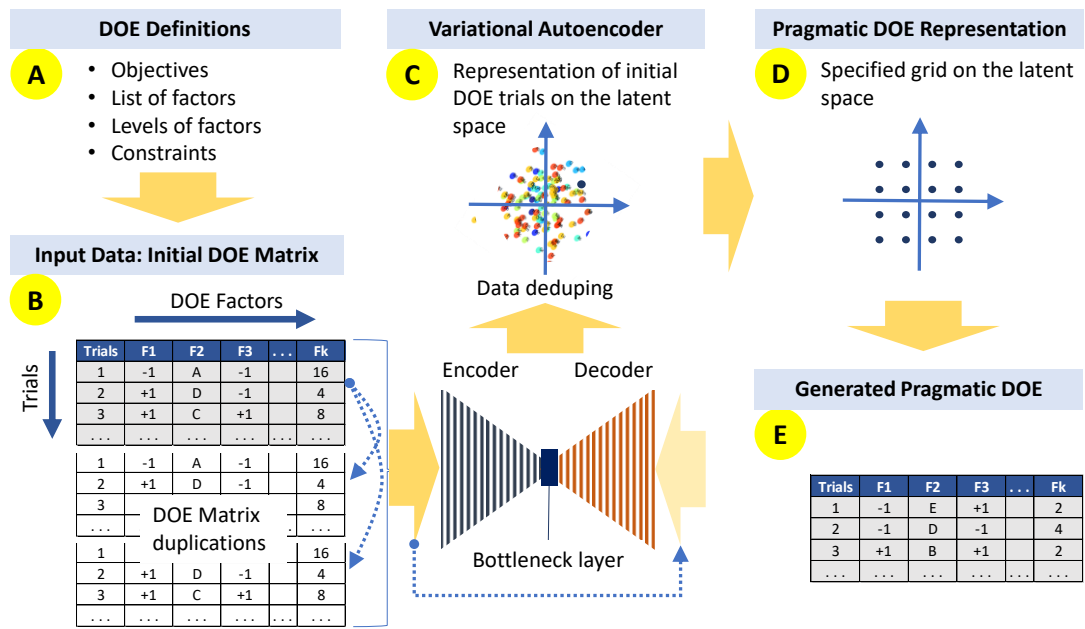

Figure 1. Process flow of designing complex experiments by applying β-VAE

The initial DOE is presented as the dataset that has the number of columns equalling the number of factors, and number of rows equalling the number of trials (Figure 1, B). The described input data has a number of challenges considering the usage of machine learning algorithms: discrete data and the number of observations that equals the number of trials. The latest poses a challenge to the fitting of autoencoders that commonly utilizes a split between training and testing datasets. To overcome these issues, simple duplications of input data have been applied (Glushkovsky, 2020). A common approach to improve robustness of the modeling process and to prevent overfitting involves the addition of Gaussian normally distributed random noise ~ $\alpha \cdot N(0,1)$ (Guozhong An, 1996). It turned out that β-VAE could be run against the duplicated dataset providing good convergence of training and testing results, decent accuracy of decoded data against input data, and that adding Gaussian noise has no noticeable effects on the robustness of the obtained results.

Duplication of the initial input matrix may be an unnecessary step when dealing with designs that have a high number of levels for many factors. It is, however, a subject for additional validation.

The next step after input data preparation is fitting of β-VAE (Figure 1, C). One of the major β-VAE hyperparameters that need to be set is the number of latent space dimensions. The representations may be set on different spaces: as low as one-dimensional, 2D, 3D, or even higher. The one-dimensional case loses isotropic property of the multivariate posterior distribution, and, therefore, cannot support orthogonality of designs. The higher the dimensionality of the latent space, the smaller the loss of the encoding-decoding propagation; but this significantly challenges interpretations. Regarding visualization and interpretability, the paper is limited to 2D representations.

Autoencoding should be tuned to provide acceptable accuracy of the encoding-decoding while forcing distribution on the latent space to be close to isotropic normal 2D distribution.

After β-VAE fitting, the next step is the specification of a grid that covers the latent space (Figure 1, D). When specifying a grid, two properties should be considered: symmetry and coverage of the space that enforces orthogonality and balance of the G-DOE. In contrast, designs that are based on randomly selected trials from the initial design have no enforcement of these properties.

Two symmetrical grids are considered in the paper: square which is suitable for a uniformed latent space and polar reflecting isotropic property of 2D normal distribution. Of course, other grids can be specified. In addition, two analytical approaches have been discussed supporting specification of grids: clustering that aggregates points on the latent space, and a gradient metric of factor levels that can be used to split the latent space into homogeneous segments with respect to the levels of factors.

The last step in the presented process flow is the generation of pragmatic DOE by decoding specified grids (Figure 1, E). The generated design has a distinct property that differs it from fractional factorial: it generates levels that may not be included in the initial design. It is an important feature affecting continuous numeric factors, while integer or binary factors are subject to rounding after decoding.

Essentially, the design of pragmatic experiments by applying β-VAE is a transformation of graphical structures. Thus, the initial design is represented as a graph on the latent space where nodes correspond to trials. Based on that representation, a grid with a reduced number of nodes that systematically covers the latent space is specified with the following generation of pragmatic DOE by decoding the nodes of the grid.

Applications of graphs in design of experiments have already been presented in numerous publications, such as squares or cubes in factors' coordinate systems, central composite designs (Box *et al*, 1978), Latin hypercube sampling (Owen, 1992; Tang, 1993), Taguchi linear L-graphs (Ross, 1996), or collections of graphical methods (Barton and Schruben, 1989; Barton, 2012, Niedz and Evens, 2016). Meanwhile in machine learning, graph representation is a rapidly developing topic (for example, Bronstein *et al*, 2021).

Concerning the proposed implementation of machine learning for design of experiments, it can be imagined as a double funnel process: across factors (columns) and then across trials (rows) of the initial DOE matrix. The first funnel propagates inherent data structure of an initial design having K factors into 2D space (reduction: K→2). The second funnel includes the specification of the grid that systematically covers the latent space with a reduced number of trials (reduction: $N_{initial} \rightarrow N_{pragmatic}$).

## 3    Properties of Initial DOE Representation on the Latent Space

Full factorial designs are orthogonal, all factors are balanced, and all trials are unbiased among each other.

On the other hand, beta autoencoding supports (1) orthogonality of the latent space dimensions, (2) isotropic multivariate normal distribution of the representation on the latent space, (3) disentanglement of the latent space representation by levels of factors, and (4) generation of trials by decoding the latent space points. It means that encoding of a full factorial design should have quite symmetrical representation and ideally, should have isotropic normal distribution on the latent space. These properties of the representation can control training of β-VAE and tune it, if required.

Beta hyperparameter is a primary control over the distribution of the representation on the latent space: the higher the beta, the more emphasized is the regularization by Kullback–Leibler divergence, and the closer the posterior distribution is to isotropic normal distribution. Of course, tuning of other hyperparameters may be required as well.

Isotropic normally distributed 2D latent space can be transformed into uniform. It allows for better visualization and interpretability of the distribution of points on the transformed latent space. In the paper, uniformed latent space means transformation of the original isotropic normally distributed space by applying normal cumulative distribution function for both axes: $(-\infty; +\infty) \rightarrow (0; 1)$.

The variational autoencoder tends to place records with similar input values as neighboring points on the latent space. It means that high density areas on the latent space represent quite similar input records while low density areas represent distinct records or even outliers.

Concerning the representation of trials, the density distribution indicates how well the design is balanced: the more uniform the density distribution is on the transformed latent space, the more balanced is the DOE. It can be visualized by plotting density contour charts.

## 4 Experimentation with Classical Full Factorial $2^4$ Design

To illustrate the application of β-VAE for DOE, let us consider the classical full factorial design having only four factors and two levels for each factor (Figure 2, a). Of course, this example has no practical value to be represented on the latent space since it is a well-known and widely used DOE, as well as its' fractional factorial $2^{4-1}$ design (Box *et al*, 1978). Nevertheless, it is quite intriguing to run β-VAE against this very lean example having only four factors and sixteen trials when commonly machine learning assumes a lot of data.

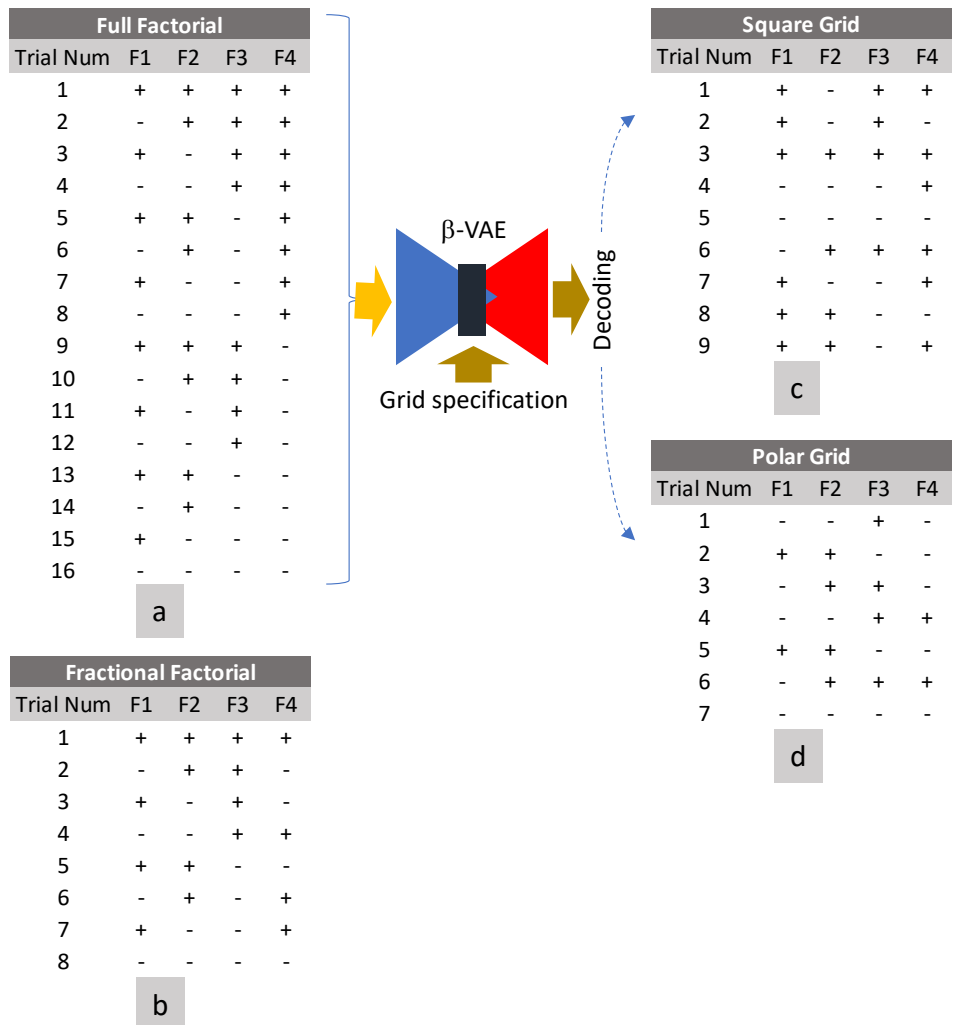

Figure 2. Full factorial $2^4$ design (a), fractional factorial $2^{4-1}$ (b), and G-DOE by decoding square (c) and polar (d) grids

Results of training β-VAE for full factorial $2^4$ design is shown in Figure 3. Despite sparse binary data and a very small size of the input matrix (16, 4), the representation of that design has quite symmetrical coverage of the latent space and reveals a very impressive disentanglement by levels of factors.

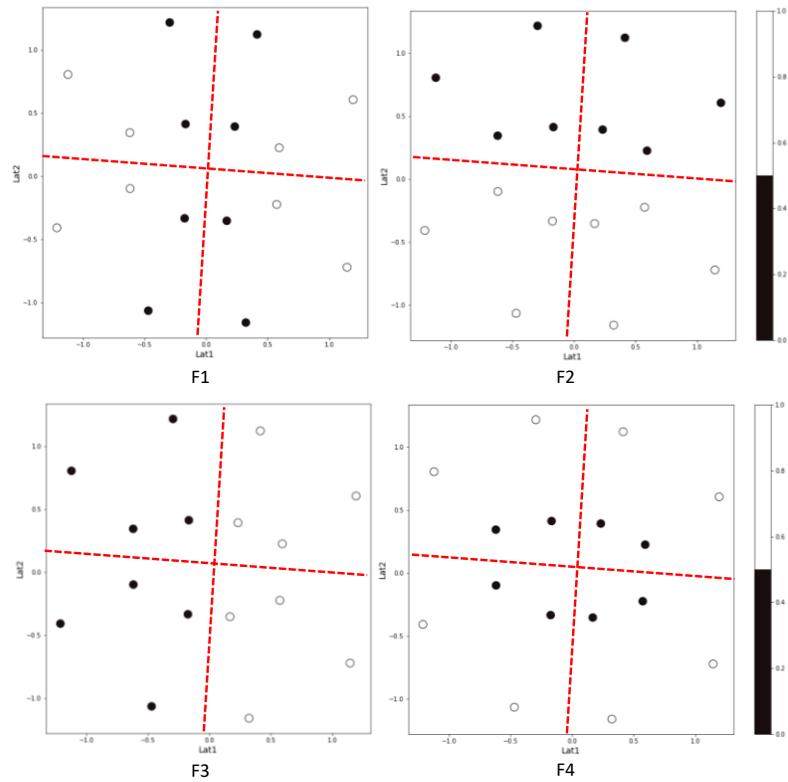

Figure 3. Representation of the classical full factorial $2^4$ design of four factors with two-levels each (N=16 trials)

Furthermore, two orthogonal axes of symmetry can be observed in Figure 3 that are represented by two red dashed lines. Taking into consideration that posterior distribution on the latent space is forced toward isometric 2D normal distribution, it is expected that these symmetry axes will not be necessarily aligned with the latent space coordinates.

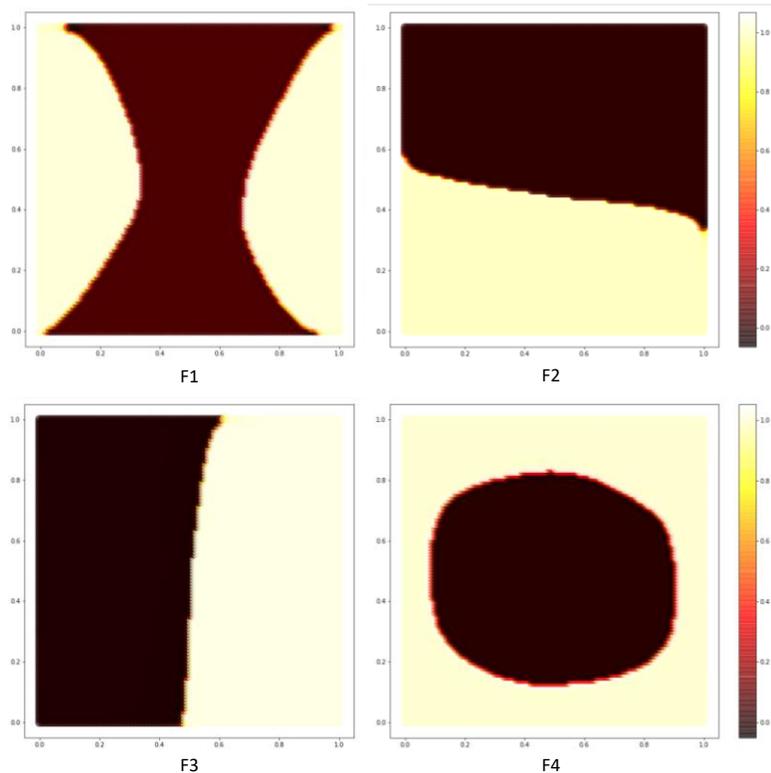

Figure 4. Illustration of disentanglement and continuous representation of factors on the uniformed latent space with decoded values

It should be noted that fitting of autoencoders is a stochastic process and there are variances of representations from run to run. Considering the obtained results, it may require tuning the model's hyperparameters or simply replicating runs. The figures of the latent space representations shown in the paper are just examples of some training outcomes.

Details concerning the applied neural networks, architectures, and finalized hyperparameters for all models used in the paper are provided in the Appendix. In addition, results shown here and later in the paper are based on the test datasets.

To illustrate the generative property of the trained β-VAE, decoding on a 100x100 grid has been applied (Figure 4). It illustrates strong disentanglement for all four factors and continuous representation on the uniformed latent space. Remarkably, the generated shapes can be associated with hyperbolic (F1), Euclidean (F2 and F3), and elliptical (F4) projections.

Having representation of full factorial design on the latent space, it is possible to generate DOE based on the defined number of trials and specified grid. The generation of DOE is decoding of the specified grid by β-VAE that produces levels of factors. Examples of two grids are presented in Figure 5: the square 3x3 (a), and the polar (b). The latter has two radial and three angular coordinates plus the central point, i.e., a total of seven points (2x3+1).

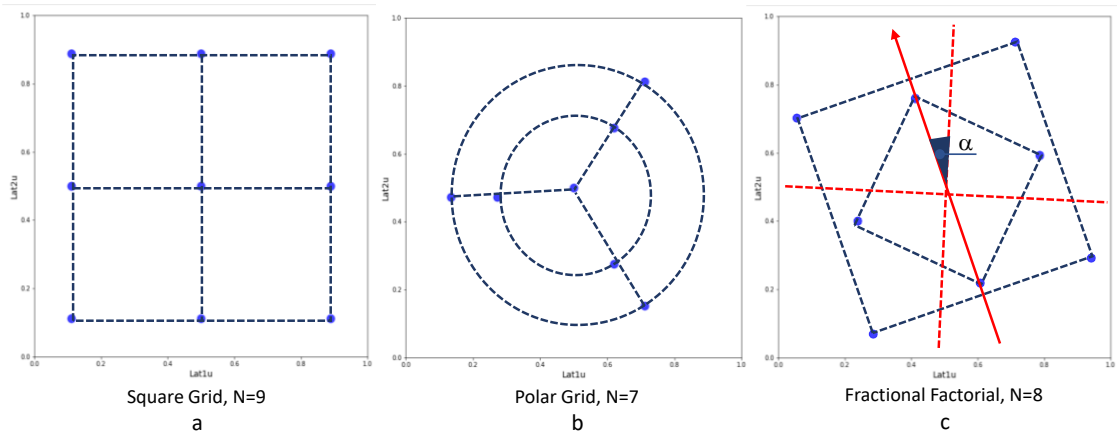

Figure 5. Examples of grids on the latent space

Certainly, other grids may be considered, but it makes sense to limit them to be symmetrical and to have close to uniform coverage of the transformed latent space.

When detecting symmetry axes on latent space representation (Figure 3, red dashed lines), it makes sense to rotate grids to some degree to avoid cases of confounded factors. Both specified grids (Figure 5, a and b) have not been subject to rotations since the symmetry axes are not aligned with the latent space coordinates.

Results of decoding both grids are shown in Figure 2, c and d. It can be observed that for both generated designs, there are no confounded factors and all of them have two levels. However, factors are not balanced and not orthogonal. It is not a surprised outcome considering that there are only sixteen points representing initial full factorial design on the latent space and the specified grids with seven and nine trials are not fractional.

These findings show the limitation of stochastic unsupervised machine learning, and it points out that supervision is still required to assess the obtained results and to tune the process, if required.

For comparison, let us consider fractional factorial design. The classical designs are carefully structured to ensure orthogonality, balance, and required resolution (Box *et al*, 1978; Montgomery, 2012). The fractional factorial $2^{4-1}$ matrix is shown in Figure 2, b, and the corresponding grid on the latent space is shown in Figure 5, c. The grid consists of two squares with 90° relative rotation of each other. It can be noticed that the grid itself is symmetrical but rotated from one symmetry axis of the latent space on about α = π/8 (22.5° degrees) counterclockwise. The clockwise rotation of the grid on about π/8 degree clockwise also generates fractional factorial design. That rotation of π/8 corresponds to the number of trials N=8.

Disentanglement of levels of four factors by β-VAE shown in Figure 4 splits the latent space into sixteen segments that represent trials of the full factorial $2^4$ design. Thus, factor F1 forms the saddle area, factor F2 splits the space vertically in half, factor F3 splits the space horizontally in half, and factor F4 creates the circular area. Schematically, these sixteen segments are presented in Figure 6.

Eight red segments are associated with the fractional factorial design (Figure 2, b). The white segments form another version of the fractional factorial $2^{4-1}$ design. Rotations of 45° switch between two versions of $2^{4-1}$ fractional factorial designs. It can be noted that the red as well as the white segments have no adjacent borders.

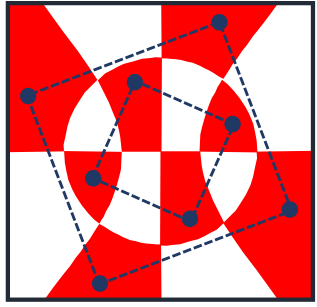

Figure 6. Schematic representation of full factorial $2^4$ trials on the latent space, where blue points represent fractional factorial $2^{4-1}$ design

# 5 Illustrative Example of a Complex Experiment

To demonstrate the ability of β-VAE to support the design of complex experiments, let us consider tuning of a convolutional neural network (CNN) to recognize handwritten digits. The input data includes 5,000 MNIST images of 0-9 handwritten digits (LeCun and Cortes, 2010). The applied typical CNN that consists of five hidden layers is described in the Appendix.

Nine hyperparameters of the CNN with mixed levels varying from two to five have been selected for tuning to maximize the accuracy of the CNN classification. As such, the accuracy of the CNN classification has been defined as a response variable. The initial full factorial design has 7,200 trials.

Details of the initial design of experiment are shown in Table 1.

| Factor ID | Name | Layer | Type | Parameter | Number of levels | Levels |
|---|---|---|---|---|---|---|
| 1 | n1 | 1 | Conv2D | Number of filters | 5 | 8, 32, 128, 512, 2048 |
| 2 | k1 | 1 | Conv2D | Kernel size | 3 | 3, 5, 7 |
| 3 | a1 | 1 | Conv2D | Activation function | 2 | relu, tanh |
| 4 | p1 | 2 | MaxPool2D | Pool size | 2 | 2, 4 |
| 5 | n2 | 3 | Conv2D | Number of filters | 5 | 8, 32, 128, 512, 2048 |
| 6 | k2 | 3 | Conv2D | Kernel size | 3 | 3, 5, 7 |
| 7 | a2 | 3 | Conv2D | Activation function | 2 | relu, tanh |
| 8 | p2 | 4 | MaxPool2D | Pool size | 2 | 2, 4 |
| 9 | d | 5 | Dropout | Dropout rate | 2 | 0.25, 0.50 |

Table 1. Illustrative DOE example

To demonstrate representation of the constrained designs by β-VAE, let us introduce the following conditions limiting CNN architecture to be a "funnel" type:
- the number of first layer filters should be greater than the number of filters of the second layer ($n1>n2$)
- the kernel size of the first layer should be greater or equal to the kernel size of the second layer ($k1 \geq k2$)

It should be noted that these constraints are introduced for illustrative purposes only.

The first constraint essentially reduces the number of levels for $n1$ and $n2$ from five to four. Also, these conditions mean dependencies between factors that impact orthogonality and balance of the design. After filtering out violated trials, the initial DOE includes 1,920 trials.

Considering the stochastic process of CNN training, three replications have been completed for all trials and dual response methodology has been applied (Vining and Myers, 1990). Based on estimations of the mean and the standard deviation for each trial, the lower confidence limits (LCL) have been calculated at a 90% confidence level as a conservative measure of the achieved accuracy (Glushkovsky, 2018).

The example is for illustrative purposes, which is not intended to provide the optimized solution of CNN tuning for MNIST recognition problem. Thus, the data included in the example is a small subset of available examples (LeCun and Cortes, 2010), where other architectures of neural networks and hyperparameters are available, and the applied CNN training process is very lean (see Appendix for more details).

## 5.1 Disentanglement of Levels of Factors in the Latent Space Representation

The encoded representation of 1.920 trials of initial DOE on the latent space is shown in Figure 7. To linearize levels of the number of filters *n1* and *n2*, log transformation has been applied. Categories of the activation functions *a1* and *a2* have been one-hot encoded. The applied β-VAE is based on a sequential deep learning algorithm that is described in the Appendix.

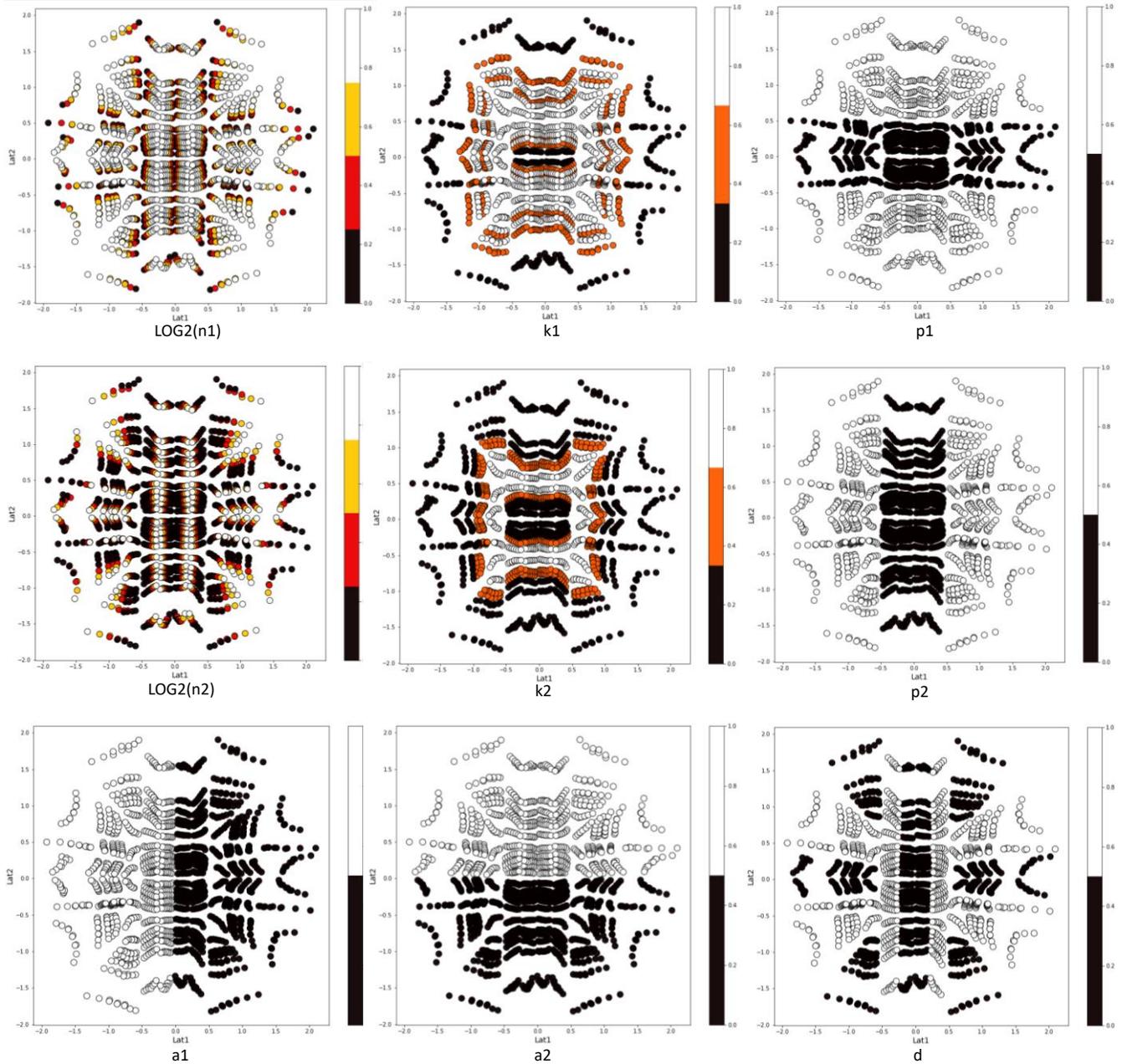

Figure 7. Unsupervised representations of factors of the initial constrained DOE on the 2D latent space (N=1,920)

The shown representations of initial DOE trials on the latent space reveal notable patterns for all nine factors. This is an especially remarkable result acknowledging that initial DOE was subject to two constraints that disrupt orthogonality and balance that has been embedded in the full factorial design of 7,200 trials.

As mentioned earlier, β-VAE training is a stochastic process, and Figure 7 provides an example where significant variance of representations from run to run should be expected.

Autoencoding usually requires tuning hyperparameters, such as the number of duplications, the beta parameter, and the number of nodes. Tuning the beta parameter reveals that small values, such as the defined β=0.3, provide meaningful disentanglement of factor levels and quite isotropic distribution while preserving good decoding accuracy.

The constrained initial design (i.e., full factorial after the elimination of trials that violate applicable constraints) is not necessarily balanced and orthogonal. However, its representation on the latent space by β-VAE still tends to be isotropic normally distributed on the latent space. Generating a pragmatic design, by selecting a symmetrical grid that covers the latent space, will drive these properties towards orthogonality and balance of the G-DOE.

### 5.2 Continuous Representations of Factors on the Latent Space and Generation of Trials

Generative VAE can produce outputs for any location on the latent space (Kingma and Welling, 2014). Examples of such generation are shown in Figure 8 by applying a high-resolution grid (100x100) on the uniformed latent space followed by decoding. It provides visualization of each factor mapping on the latent space as color coded points. Clearly, the distributions of factors levels have very distinct patterns on the latent space.

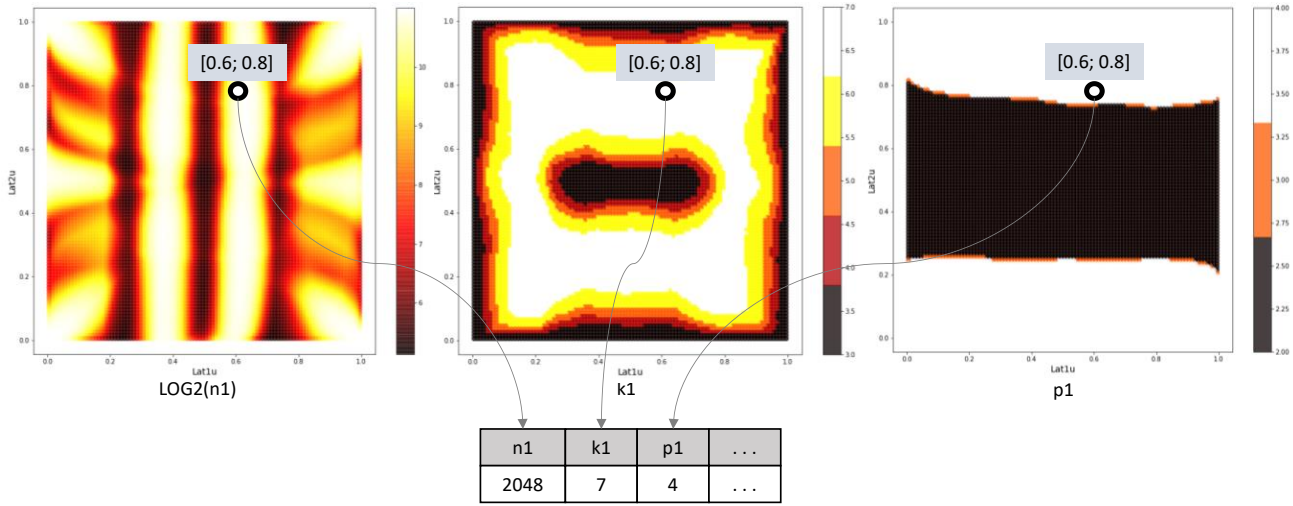

Figure 8. Examples of the decoding of the factors' values on a 100x100 grid illustrating disentanglement and continuous properties of representations on the uniformed latent space

Example of decoding one point with transformed coordinates [0.6; 0.8] is provided for the first three factors.

### 5.3 Propagation of Constraints into the Latent Space Representation

Decoding of factors' levels on a 100x100 grid, it turned out that representation on the latent space substantially preserves constrained conditions that have been applied to full factorial DOE. Only six points out of 10,000 have been found violating these constraints (Figure 9). It means that constraints are embedded in the representation and will be reflected in the generated pragmatic design.

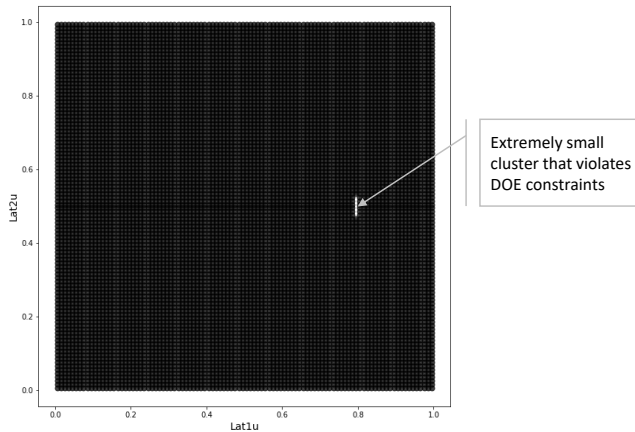

Figure 9. Representation on the latent space has an extremely limited number of trials that violate constraint conditions

Noting this very important outcome, it is still advisable to check that all trials are compliant to the constraint conditions when generating a pragmatic DOE and to tune the generation process by changing either the number of trials or the grid pattern, if necessary.

## 5.4 Specification of Grids on the Latent Space to Generate Experiments

Even though machine learning is a stochastic process and representations on the latent space are not exactly reproducible from run to run, the disentanglement, isotropic posterior distribution, and generative properties, allow for the generation of a pragmatic G-DOE.

Specifying the number of trials of the pragmatic G-DOE, the balance between design objectives, including design resolution, versus economical, operational, and timing issues should be taken into consideration.

For illustrative purposes, let us define pragmatic experiments with a very significant drop in the number of trials from the initial design of 1,920 trials. The following three types of experiments have been considered (Figure 10): two symmetrical grids that systematically cover the latent space: the square (64 trials, i.e., reduction by 30 times) and the polar (40 trials, i.e., reduction by 48 times), and random selection of trials from the initial design (64 trials). It should be noted that the square grid is better fitted to the uniformed space, while the polar grid is more suitable for the original isotropic 2D normally distributed latent space.

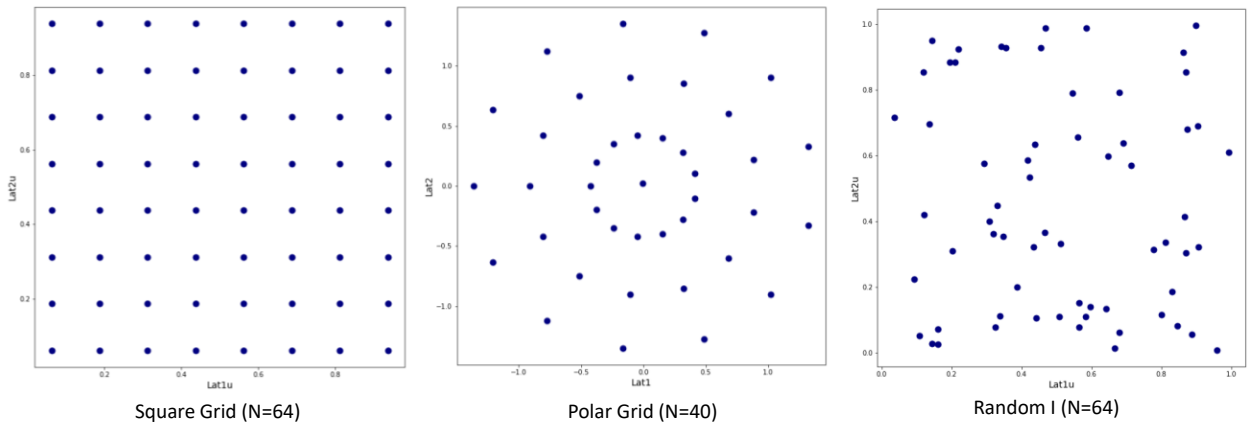

Square Grid (N=64)  Polar Grid (N=40)  Random I (N=64)

Figure 10. Examples of different grids applied on the latent space to generate pragmatic designs with the defined number of trials

To increase confidence in analysing results of the random design, it was replicated three times using different random seeds.

Random design means a simple subset from initial trials that does not require autoencoding. In this case, however, properties of this design type are dependent on chance, while the application of β-VAE provides a more systematic and controllable approach of the initial design reduction.

When specifying grids, it is important to decode trials and to check that G-DOE has no confounded factors, that each factor has a reasonable coverage of levels, all trials are not violating constraints, and only unique trials are included. The latest should not be confused with the replication of trials, which is a separate issue focusing on the confidence of response estimation.

In the event that the decoded trials are not satisfying the above-mentioned conditions, modification of the grid is required. This interactive process precedes actual experiments and ensures that concerning issues are addressed during the preparation stage.

## 5.5 Density of the Latent Space Representation as a Reflection of the Balance of Design

Distribution of trials on the latent space provides important insights when designing experiments. The β-VAE tends to represent close input records as neighboring points on the latent space. It means that the more uniform the distribution is on the transformed latent space, the more balanced the G-DOE will be.

Visualizations of density distributions of different experiments are shown in Figure 11. They indicate that full and both grids (square and polar) have quite symmetrical distributions and substantial uniformity. On the other side, all three random designs clearly have areas with significantly high or low concentrations of trials on the latent space, and, therefore, poorly balanced designs.

Observing density distributions allows for the control of designs by specifying different grids and the number of trials. Thus, taking a closer look on the polar grid distribution suggests that it can be improved. For example, by either increasing the number of points on the two outer circles, or by regrouping points between the inner and outer circles while keeping the total number unchanged.

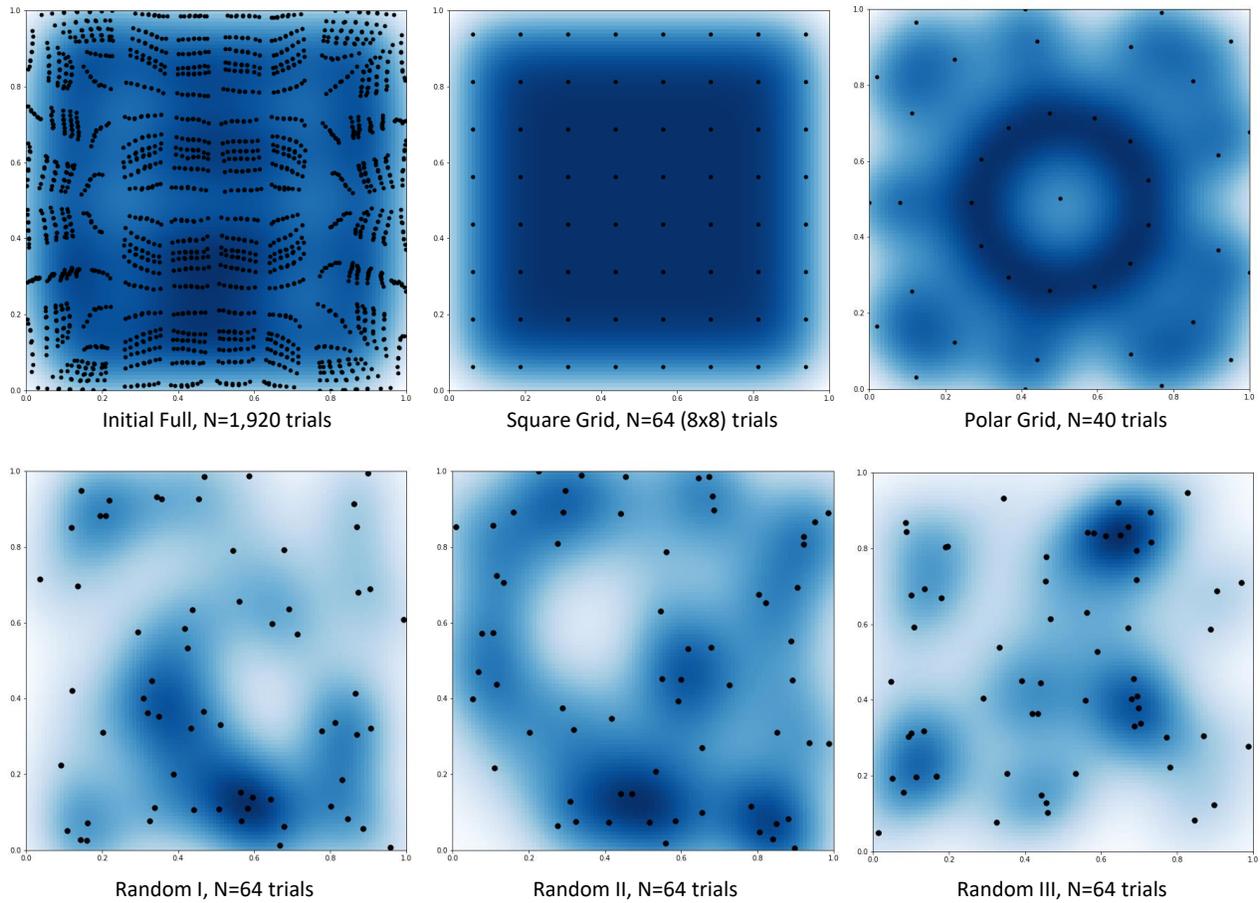

Figure 11. Densities of trials on the latent space for different types of experiments

### 5.6 Visualization and Interpolation of Response on the Latent Space

Representations of responses for executed experiments are shown in Figure 12. Such representations become possible by dramatically reducing dimensionality of the DOE matrix by applying an autoencoder. In our illustrative example, the initial nine factors have been embedded into a 2D space.

For better visualization, linear interpolation on a 100x100 grid has been applied for all six executed experiments. Polynomial interpolations or other techniques, such as spline smoothing, can be applied to map estimated responses on the high-resolution grids on the latent space. However, it should be noted that the linear interpolation is quite conservative and prevents overfitting.

The linearly interpolated accuracy metric plotted on the uniformed latent space for the initial design of 1,920 trials has a very complicated surface with some patterns and symmetries. It means that factors have definite impacts on the response (Figure 7). However, there is a very strong non-linear behaviour of the response across the latent space and there are multiple local maximums. It means that the design of pragmatic experiments and subsequent analysis of results will be very challenging tasks.

Some response patterns and symmetries, but with less intense irregularities, can be observed for the square grid design. The polar grid has a smaller number of trials, but even then, some indication of an existing pattern with some symmetries can be seen. Random designs have irregular coverage of the latent space, where some areas have almost redundant trials, while other areas have no coverage at all. It leads to an unbalanced design representing different levels of factors and impacts orthogonality. As a result, shown heatmaps of random designs have no signs of any organized patterns or symmetries of the response.

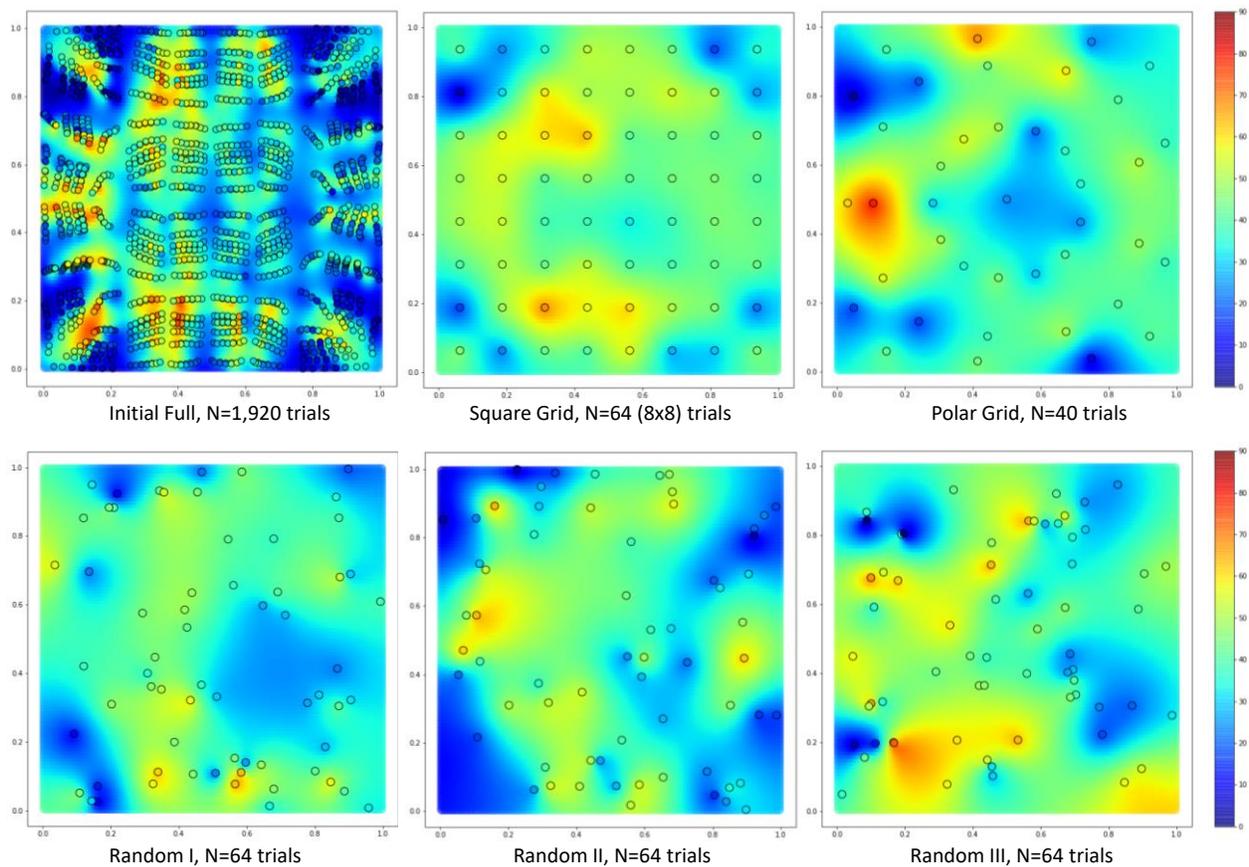

Figure 12. Interpolated heatmaps of the accuracy metric on the latent space for different types of experiments, where dots represent trials

## 5.7 Optimized Results for Different Grids

The major objective of the illustrative example is finding the optimized setting of the CNN hyperparameters that maximizes its' classification accuracy. An optimal point can be defined based on the selection of the best trial of the performed G-DOE or by modeling the response as a function of factors first and then solving for the optimal point. The first approach is based on an actual but stand-alone result, while the second approach is based on the modeled estimations that incorporates information concerning other trials as well. Non-linear regressions, gradient boosting, or neural networks can be considered for modeling assuming a large number of G-DOE trials, small number of factors, and a smoothed response surface. To prevent overfitting while incorporating information of the neighboring points, the linear interpolation on a high-resolution 100x100 grid on the latent space has been applied in the described example. The results of the accuracy metrics (LCL) are shown in Table 2 for the executed designs. The table also includes decoded levels of the optimized CNN hyperparameters and coordinates of the optimal points on the latent space.

| | CNN Hyperparameters | | | | | | | | | | | LCL of Accuracy Metrics | | |
|---|---|---|---|---|---|---|---|---|---|---|---|---|---|---|
| Design | n1 | k1 | p1 | a1 | n2 | k2 | p2 | a2 | d | lat1u | lat2u | Executed Trials | Interpolated | Valididated |
| Initial Full | 1322 | 7 | 4 | relu | 81 | 6 | 4 | tanh | 0.25 | 0.16 | 0.12 | 98.8% | 98.7% | 98.6% |
| Grid | 796 | 7 | 4 | relu | 25 | 6 | 2 | tanh | 0.50 | 0.32 | 0.19 | 98.5% | 98.5% | 98.4% |
| Polar | 1579 | 7 | 2 | relu | 168 | 5 | 4 | tanh | 0.25 | 0.11 | 0.49 | 98.7% | 98.7% | 98.8% |
| Random I | 1544 | 5 | 4 | tanh | 121 | 3 | 2 | tanh | 0.25 | 0.58 | 0.09 | 98.4% | 98.4% | 98.3% |
| Random II | 391 | 7 | 2 | relu | 19 | 5 | 4 | relu | 0.25 | 0.12 | 0.58 | 98.4% | 98.4% | 97.8% |
| Random III | 720 | 7 | 4 | relu | 36 | 7 | 4 | tanh | 0.50 | 0.18 | 0.20 | 98.6% | 98.6% | 97.9% |

Table 2. Optimal points for the executed experiments

It can be observed that the optimal points of the interpolated results are very close to the best trials of the executed experiments. The best results have been achieved by the initial and polar grid designs, followed by the square grid. Out of the three random designs, two underperformed.

To validate the obtained results, the trials have been replicated three times at optimal settings based on interpolations. It turns out that two random designs (II and III) have significant deviations between the interpolated and actual results (Table 2, red cells). It is caused by unsystematic trial locations of the random designs on the latent space that affect interpolation.

To provide visualization of the obtained results, optimal points are shown on the latent space with a heatmap of the accuracy metrics (LCL) of the initial DOE as a benchmark (Figure 13).

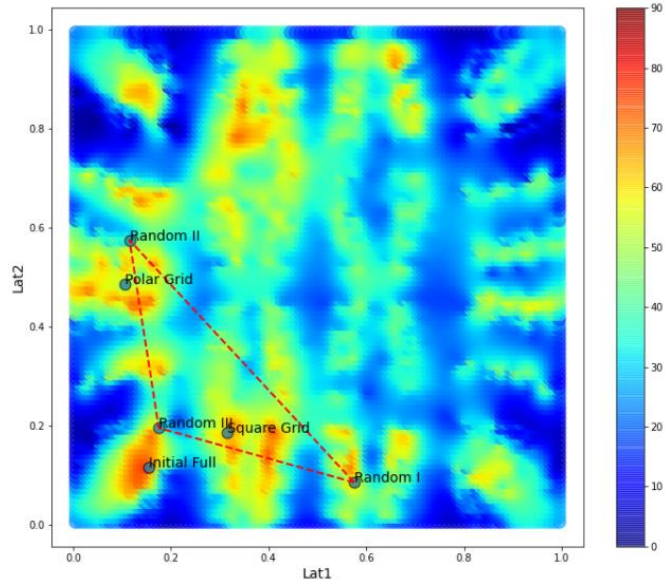

Figure 13. Optimal points on the latent space for different types of experiments

As mentioned earlier, the heatmap of the interpolated LCL accuracy metric of the constrained full factorial DOE has multiple local maximums. It is expected that the different designs capture various locations as the maximum points. Nevertheless, the differences between the validated accuracy metrics for the designs based on the systematic grids and the initial DOE, as a benchmark, are not very strong, only +/-0.2% (Table 2). For comparison, the triangle, which links replications of the random designs, indicates a high spread of optimal points (Figure 13) and high variance of the obtained results for that type of design (Table 2).

## 5.8    Evaluation of the Importance of Factors by Supervised Machine Learning

To address the importance of factors with respect to their impact on the response, supervised machine learning has been employed to model the response as a function of factors. After models have been trained for different types of experiments shown in Figure 11, the random permutations of each factor have been applied observing increases of model loss functions (Breiman, 2001; Fisher *et al*, 2018).

Supervised sequential neural networks have been trained against the obtained accuracy of CNN by applying different tuning of nine hyperparameters according to the performed designs. A description of the supervised sequential model is provided in the Appendix.

It should be noted that having quite a low number of observations (64 for square grid and random, and 40 for polar grid) and nine inputs, there is a high risk of the supervised models to be overfitted even when applying regularization terms. This risk, however, affects the predictive power of these models against new data. In our case, the models are used to evaluate the importance of factors and not for prediction.

To increase the confidence of the obtained results, ten replications have been applied for random permutations for each factor. In addition, duplications of input data have been performed to support the convergence of fitting and to split the data between training and testing datasets (see Appendix).

Despite the concern mentioned above, it turns out that the developed models provide fairly consistent results regarding the importance of factors, where the initial design (1,920 observations) has been used as a benchmark (Figure 14). It is not a surprise that the top important hyperparameters of the convolutional neural network are the number of convolutional filters (*n1* and *n2*).

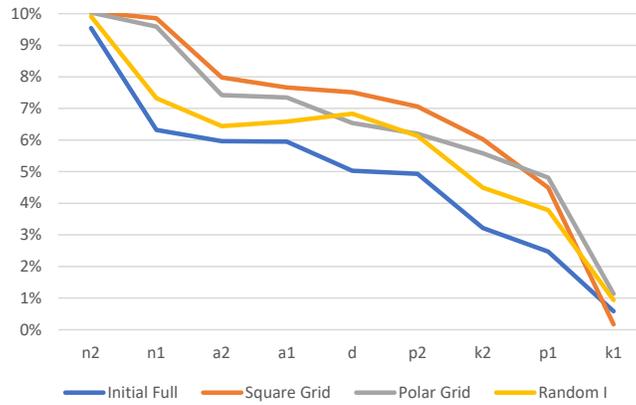

Figure 14. Importance of factors by relative loss increase given their permutations

The square and polar grids completely match the initial full design in rank of importance for all factors, while the random design has some deviation.

## 6     Utilization of Clustering to Create Grids on the Latent Space

Since trials with close levels of factors are encoded in close proximity on the latent space, there is the possibility to aggregate trials into a defined number of clusters.

Merging clustering and design of experiment topics are considered in numerous publications. For example, utilization of cluster analysis to generate DOE where factors are interdependent is provided in (Zemroch, 1989), implications of data clustering in experiments are discussed in (Nahum-Shani et al. 2018), and a comparative study on clustering methods by using DOE is given in (Puggina Bianchesi *et al*, 2019).

Clustering is an unsupervised method that can be set using different algorithms and criteria of aggregation, such as distances or densities.

For illustration proposes, two of the most popular clustering methods have been applied: Ward's and K-Means. Input data in both examples includes the initial DOE representation of 1,920 trials by the 2D latent vector. Since the selected clustering aggregations are based on the distance related metrices, the transformation of the 2D normal latent space into a uniformed one is required prior to clustering.

Mappings of clusters for both methods are presented in Figure 15.

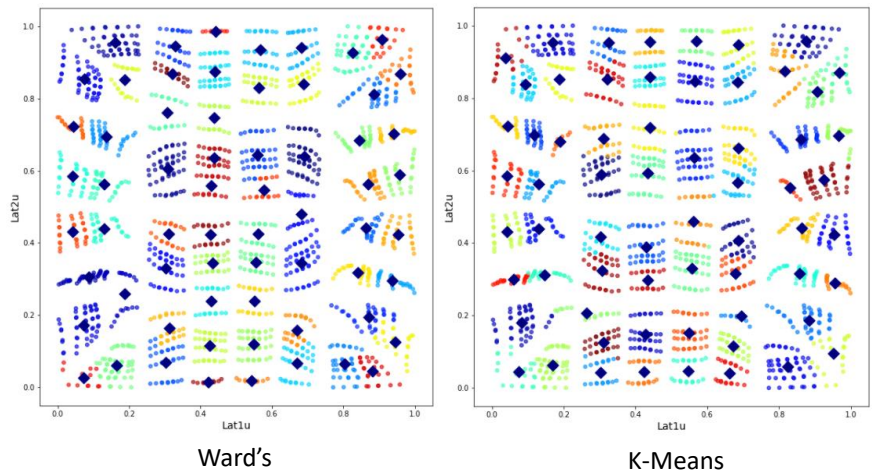

Ward's                        K-Means

Figure 15. Mappings of clusters on the latent space, where the initial design trials are color coded according to their aggregations by clusters and navy diamond points represent centroids of clusters (N=64)

It can be noted that both clustering methods provide slightly different allocations of the centroids and that density distributions of centroids are less uniform when compared to the square 8x8 grid (Figure 11). Pragmatic G-DOE based on Ward's clustering method provides a marginally suboptimal result compared to the square grid. The lower confidence limit of the accuracy metric for Ward's clustering has been estimated as 98.4% versus 98.5% for the square grid (Table 2). Therefore, the clustering approach to create grids can be seen as an interactive analytical tool that may apply different methods and tunings.

## 7    Gradient Metric of Factor Levels on the Latent Space

Decoding of factor levels on the high-resolution grid of the latent space, as shown in Figure 8, can be used to calculate gradients along axes. Aggregating these gradients for all factors and latent dimensions, it allows for gradient metric mapping of the initial design on the latent space:

$$Gradient\_Metric(x,y)=\sum_{i=1}^{9}\sum_{j=1}^{2}\left|\frac{\Delta F_i(x,y)}{\Delta L_j}\right| \quad (1)$$

where $F_i$ is the decoded level of the $i$-factor that has been normalized to the range [0,1]; $L_j$ is the transformed $j$-axis of the uniformed latent space, and (x,y) are grid coordinates along the latent space axes.

Summarization has been applied for aggregation in the formula (1), however, different approaches, such as a MAX can be set as well. Also, the formula can be generalized for higher dimensionality of the latent space and a various number of inputs.

This metric provides segmentation of the latent space by areas of small changes of factors' levels that are separated by borders representing high-level changes (Figure 16). It can be seen as a well-known first order image derivative.

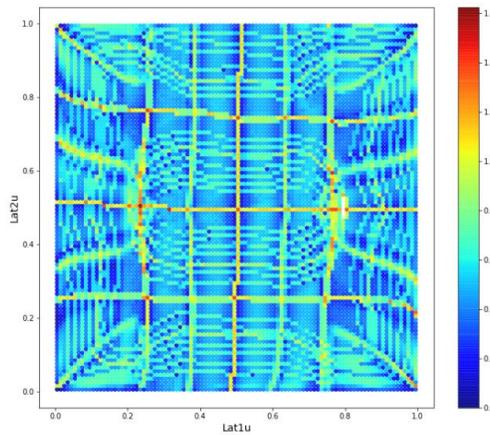

Figure 16. Mapping of the gradient metric on the latent space

Application of the described gradient metric allows for interactive control on the number of trials and grid specifications. To simplify the control, borders can be highlighted by applying a threshold limit to the gradient metric. For example, considering the map in Figure 17, a, it suggests that the 8x8 square grid has reasonable coverage of areas with low gradients. In contrast, a substantial portion of Ward's clustering centroids are located on the border lines (Figure 17, b), while K-Means clustering centroids have slightly better positions (Figure 17, c). Of course, different threshold limits followed by an adjustment of point locations within segments can be applied.

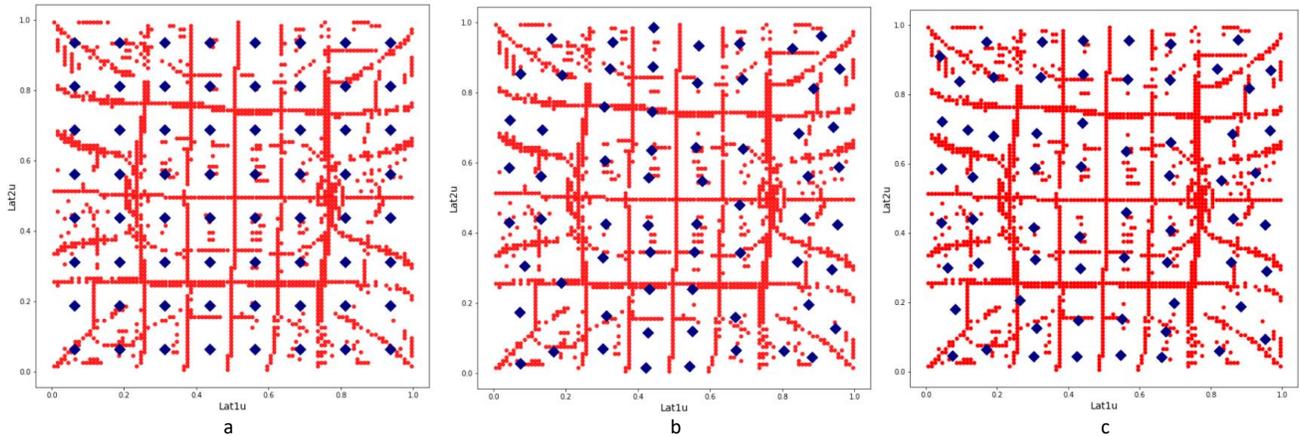

Figure 17. Visualization of gradient metric borders (red points) by applying a threshold limit of 0.5, where blue diamond points represent:
(a) an 8x8 square grid, (b) Ward's and (c) K-Means clustering centroids

# 9    Conclusions

The research paper discusses the application of unsupervised autoencoders as a supportive tool for designing complex experiments.

The process flow includes five steps: (1) definition of DOE attributes (objectives, list of factors, levels of factors, and constraints), (2) preparation of the initial full factorial design after filtering out unfeasible trials, (3) autoencoding that represents the initial design on the latent space, (4) specification of the number of trials in the pragmatic design and their pattern on the latent space, and (5) generation of the pragmatic design by decoding latent coordinates of the specified grid. Steps (4) and (5) can be imagined as a double funnel process: first, across factors and then, across trials of the initial DOE matrix. Essentially, the process of designing complex experiments by applying β-VAE is a transformation of graphical structures.

The variational autoencoder supports (1) orthogonality of the latent space dimensions, (2) isotropic multivariate standard normal distribution of the representation on the latent space, (3) disentanglement of the latent space representation by levels of factors, (4) propagation of the applied constraints of the initial design into the latent space, and (5) generation of trials by decoding latent space points. Generating a pragmatic design by selecting a symmetrical grid that covers the latent space will drive these properties towards orthogonality and balance of the G-DOE.

Applying autoencoding for the classical full factorial $2^4$ design that consists of only four two-level factors and sixteen trials, the representation of that design has fairly symmetrical coverage of the latent space and reveals a very impressive disentanglement by levels of factors.

It should be noted that autoencoding is a stochastic process and there are variances of representations from run to run and tuning of hyperparameters may be required.

For better visualization of the response surface on the latent space, linear interpolation on a high-resolution grid has been applied for the executed G-DOE. Polynomial interpolations or other techniques, such as spline smoothing, can be applied to map responses, but the linear interpolation is quite conservative and prevents overfitting.

The illustrative example included in the paper suggests that the systematic grids that uniformly cover transformed latent space, such as square or polar, provide more controllable results, while random selection of trials delivers poorly balanced design and leads to accidental outcomes.

Generative VAE can produce outputs for any given location which means continuous representation on the latent space. It may generate levels of continuous numeric factors that have not been included in the initial design.

To assist with the definition of the number of trials and grid specifications of the pragmatic design, density charts of the G-DOE representations on the transformed latent space have been considered. Density distributions provide indications on the balance of the design. The closer the distribution is to the uniformed one, the better the balance of the design. In addition, clustering analysis and the gradient metric of factor levels have been applied as interactive analytical tools that support decision making concerning the specification of the G-DOE. The former aggregates points on the latent space into clusters and the latter can be used to split the latent space into homogeneous segments of trials.

## Disclaimer

The paper represents the views of the author and do not necessarily reflect the views of the BMO Financial Group.

## Appendix

Architectures of the basic neural network models used in this research have been as follows.

β-VAE network (both classical full factorial $2^4$ and MNIST tuning examples):

Encoder: Flatten input with 4 (full factorial $2^4$) or 9 (MNIST tuning) channels; Dense (512, activation='relu'); Dense (32, activation='relu'); two Dense (latent_dim=2)

Decoder: Dense (32, input_dim=2, activation='relu'), Dense (512, activation='relu'), Dense (original_dim=4 (full factorial $2^4$) or 9 (MNIST tuning)), activation='sigmoid')

Loss=binary_crossentropy+beta*KL; beta=0.3; optimizer=adam; batch_size=256; epochs=500; data sets duplications: full factorial $2^4$: train x50 and test x30; MNIST tuning: train x5 and test x3

Classification convolutional neural network:

Input 784 (28x28); Conv2D (n1, kernel_size=(k1, k1), padding='same', activation=a1); MaxPool2D (pool_size=(p1,p1)) Conv2D (n2, kernel_size=(k2, k2), padding='same', activation=a2); MaxPool2D (pool_size=(p2, p2)); Flatten (); Dropout (d); Dense (num_classes=10), activation='sigmoid')

Loss= categorical_crossentropy; optimizer=adam; metric=accuracy; batch_size = 64; epochs = 10

Sequential neural network (importance evaluation model):

Dense (16, kernel_initializer='normal', activation='relu', kernel_regularizer=regularizers.l1_l2(l1=1e-5, l2=1e-4), bias_regularizer=regularizers.l2(1e-4), activity_regularizer=regularizers.l2(1e-5))

Dense (4, kernel_initializer='normal', activation='relu', kernel_regularizer=regularizers.l1_l2(l1=1e-5, l2=1e-4), bias_regularizer=regularizers.l2(1e-4), activity_regularizer=regularizers.l2(1e-5))

Dense (1, kernel_initializer='normal')

Loss='mean_squared_error'; optimizer='adam'; batch_size = 128; epochs = 100; data sets duplications: square and polar grids, random: train x50 and test x30; initial design: train x5 and test x3